\documentclass[10pt, a4paper]{article}

\usepackage[final]{lrec-coling2024} % this is the new style
\usepackage{adjustbox}
\usepackage{booktabs}
\usepackage{caption}
\usepackage{subcaption}
\usepackage{amsmath}
\usepackage{amsfonts}
\usepackage{xcolor}
\usepackage{hyperref}
\hypersetup{
    colorlinks=true,
    linkcolor=blue,
    filecolor=magenta,      
    urlcolor=black,
}

\title{Black-Box Hallucination Detection via Consistency Under the Uncertain Expression}

\name{Seongho Joo, Kyungmin Min, Jahyun Koo, Kyomin Jung}

\address{
    Seoul National University \\
    \{seonghojoo, kyungmin97, koojahyun, kjung\}@snu.ac.kr
}

\abstract{
Despite the great advancement of Language modeling in recent days, Large Langauge Models (LLMs) such as GPT3 are notorious for generating non-factual responses, so-called "hallucination" problems. Existing methods for detecting and alleviating this hallucination problem requires external resource or the internal state of LLMs, such as the output probability of each token.
Given the LLM's restricted external API availability and the limited scope of external resources, there is an urgent demand to establish the Black-Box approach as the cornerstone for effective hallucination detection.
In this work, we propose a simple black-box hallucination detection metric after the investigation of the behavior of LLMs under expression of uncertainty. 
Our comprehensive analysis reveals that LLMs generate consistent responses when they present factual responses while non-consistent responses in vice-versa. Based on the analysis, we propose an efficient black-box hallucination detection metric with the expression of uncertainty. 
The experiment demonstrates that our metric is more predictive of the factuality in model response than baselines that use internal knowledge of LLMs.
\\ \newline  }

\begin{document}

\maketitleabstract
\section{Introduction}
In recent days, Large Language Models (LLMs) have made great progress in natural language generation. Based on a vast amount of training corpus, they have shown remarkable performance in open-domain QA, abstractive summarization, and machine translation~\cite{goyal2023news, ouyang2022training, touvron2023llama}. 
%However, despite the remarkable performance, LLMs are prone to generate non-factual statements that are not grounded in the existing training corpus, which is widely known as the hallucination problem~\cite{Ji2022SurveyOH}.

To detect and mitigate the hallucination problem, many hallucination detection methods rely on external resources~\cite{Shuster2021RetrievalAR,Ji2022RHOR, Guerreiro2022OptimalTF} or require internal state of LLMs like probability per output token ~\cite{azaria2023internal, Kuhn2023SemanticUL}, 
and few black-box methods that only require the response of LLMs are proposed~\cite{Manakul2023SelfCheckGPTZB,cohen2023lm}. 
However, we often do not have access to the internal information of LLMs or external sources but only their generated output. 
Furthermore, external knowledge sources like Wikipedia may not cover sufficient knowledge to answer the question.
It is also reported that using the retriever model for question answering may worsen the performance when dealing with factual knowledge in long-tail \cite{mallen2023trust}.

Previous black-box hallucination detection methods rely on obtaining around ten LLM responses for successful detection~\cite{Manakul2023SelfCheckGPTZB}. These sampling-based methods come with significant computational costs and can pose challenges in determining the appropriate sampling temperature for detection. However, our approach necessitates just a single extra model inference.

\begin{figure}[h]
    \centering
    \includegraphics[width=\linewidth]{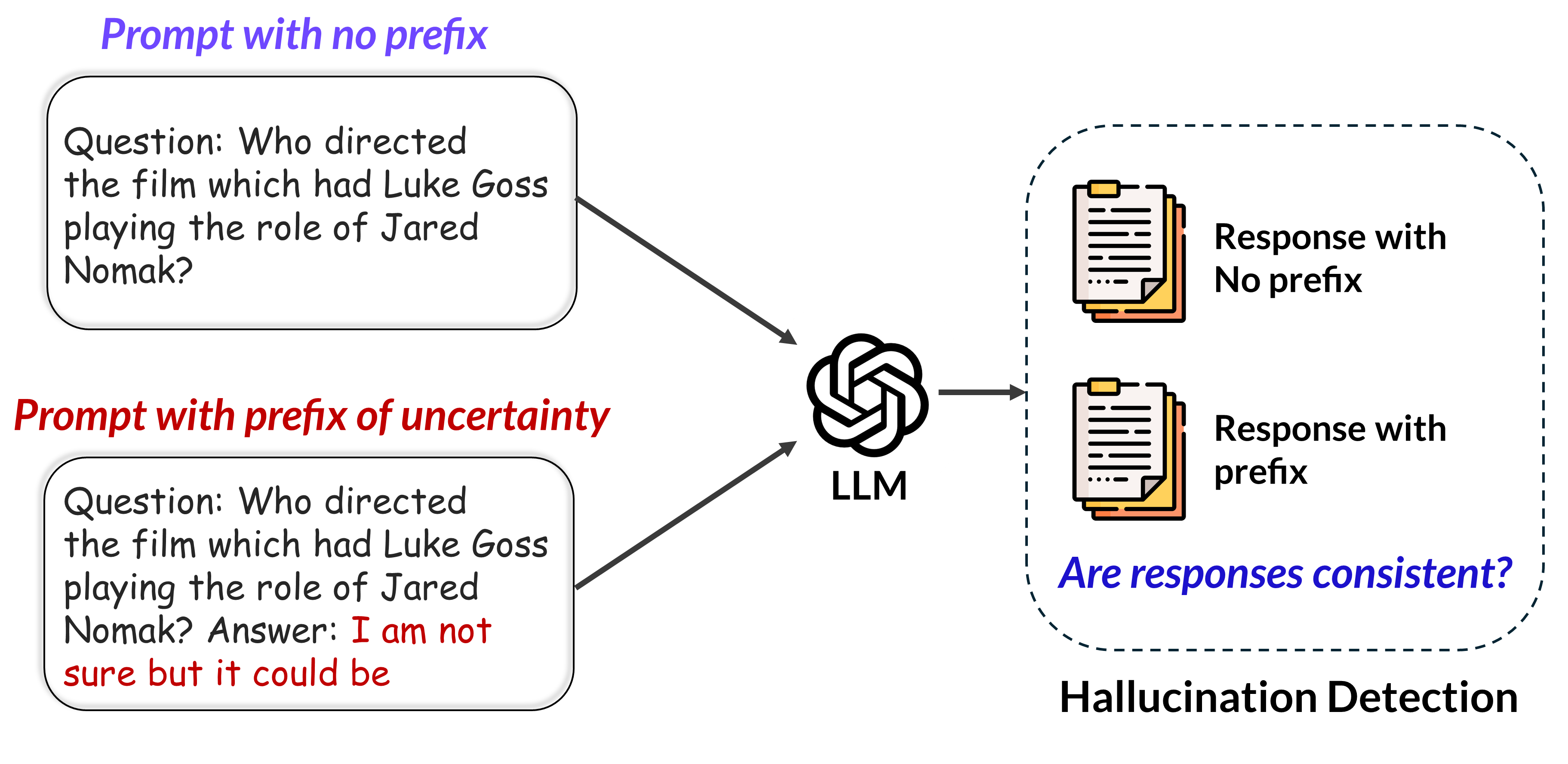}
    \caption{Example of prompt-based hallucination detection without an external source. Given the question, the expression of uncertainty is used as the prefix for the answer part. The answer before the prompt perturbation and the response with the new prompt are compared.}
    \label{figure1}
    \vskip -0.2in 
\end{figure}

In this work, we aim to develop an efficient prompt-based hallucination detection method that does not require multiple sampling of responses. We explore prompts with expressions that can \textbf{discriminate} between factual responses and non-factual responses generated by LLMs as shown in Figure \ref{figure1}.
%since 문장 다시 생각
Motivated by the fact that humans express their level of confidence when answering questions, our preliminary study examines what happens when LLMs express their level of confidence in their answers.
Therefore, we investigate the following question: "Do LLMs behave consistently or differently under expression of uncertainty when they know the answer?".  To answer this question, we gather expressions of uncertainty such as "I am not sure but it could be" and expressions of certainty, such as "It must be." These expressions serve as prompts for our investigation like Figure~\ref{figure1}. Through the experiment, we aim to answer the following question: 
\begin{itemize}
    \item Do LLMs generate consistent responses under uncertain or certain expressions?
    \item Is there a difference in terms of response accuracy or confidence of LLMs between when the model answers with uncertain expression and certain expression?
\end{itemize}
In Section \ref{preliminary}, we conduct a comprehensive analysis regarding the questions on two open-domain question answering (QA) benchmarks, HotpotQA \cite{Yang2018HotpotQAAD}, and NQ-open \cite{kwiatkowski-etal-2019-natural} in the closed-book setting. 
We evaluate the performance of GPT-3 \texttt{text-davinci-003} in a zero-shot manner and divide the (question, answer) pair into two groups. Specifically, we divide the samples into the factual group and the non-factual group based on the factuality of the original response before the prompt perturbation. 
In both datasets, there is a clear distinction between the two groups when the expression of uncertainty is used. Most samples in the factual group show consistency in the response, while samples of the non-factual group often show inconsistency in the response after the prompt is changed. In addition, we also find the number of samples that show consistency differs not much in the response between when uncertain expression and certainty are used. 

Based on our findings,  we propose a simple hallucination detection metric using only a few number of responses of LLMs in Section \ref{detection metric}. Specifically, we utilize the consistency between the response generated using the standard prompt and the response generated using a prompt that includes an uncertain expression.  Unlike previous detection metrics, our metric does not require the internal state of LLMs nor multiple responses through multiple sampling. The experiment demonstrates that our metric efficiently distinguishes factual answers compared to the baseline that requires internal knowledge of LLMs.

\section{Related Work}

\subsection{Large Language Models}

Large Language Models (LLMs) are pre-trained on massive amounts of text corpora, and this pre-training equips LLMs with powerful text generation capabilities and zero-shot abilities. \cite{brown2020language,zhang2022opt, chowdhery2022palm,touvron2023llama}.
Through instruction tuning and reinforcement learning, such as reinforcement learning from human feedback (RLHF), LLMs demonstrate better alignment with user intent, resulting in improved zero-shot abilities and in-context learning~\cite{ouyang2022training, wei2022finetuned,longpre2023flan,chung2022scaling}.
Additionally, instruction tuning enhances LLMs' proficiency in user interactions, leading to the emergence of practical models ~\cite{Liu_2023,openai2023gpt4,vicuna2023}.
However, practical LLMs need to be improved in detecting hallucinations due to the constraints of external APIs, which hinder the devise of hallucination detection methods relying on LLMs' internal states~\cite{azaria2023internal, Kuhn2023SemanticUL}.
Our work proposes a black-box hallucination detection method that distinguishes factual answers more effectively than previous methods without requiring access to the internal states of LLMs.

\subsection{Hallucination Detection of Large Language models}
One way for detecting hallucinations involves using external sources (e.g., Wikipedia) to evaluate whether the generated text contains non-factual information~\cite{thorne-etal-2018-fact,augenstein-etal-2019-multifc,Shuster2021RetrievalAR,Ji2022RHOR, Guerreiro2022OptimalTF}.
Another approach is to detect hallucinations by leveraging the internal state of LLMs.
The internal state of LLMs, such as output token-level probabilities, enables the derivation of a confidence score for assessing hallucination~\cite{mielke-etal-2022-reducing,Kadavath2022LanguageM,varshney2023stitch,xiao-wang-2021-hallucination, malinin2021uncertainty}. 
However, applying external references have limitation in real-world scenarios due to their inability to cover all factual knowledge, and obtaining LLMs' internal states is challenging due to limited API functionalities.
To address these issues, some black-box methods overcome these limitations by relying solely on samples generated by LLMs.
\citet{cohen2023lm} detect factual errors by cross-examining responses generated by two LLMs in a multi-turn interaction. \citet{Manakul2023SelfCheckGPTZB} propose that LLMs generate multiple samples and evaluate the consistency of these samples as a basis for hallucination detection. 
These methods, however, have a drawback in terms of inference cost due to using multiple samples and the challenge of identifying an appropriate sampling method for detection.
Our hallucination metric doesn't necessitate multiple responses. All we need is a single, standard response and a response that expresses certainty.
\begin{table*}[t]
    \centering
    \begin{subtable}[t]{0.48\textwidth}
        \resizebox{\textwidth}{!}{\begin{tabular}{c| c c c}
        \toprule 
            \textbf{Group} &  \textbf{Factual} & \textbf{Non-Factual} & \textbf{Total}\\
            \midrule 
             log probability &   $-0.128$ & $-0.280$ & $-0.228$ \\ 
             Entropy  & $+0.160$ & $+0.343$ & $+0.280$\\ 
             \bottomrule
        \end{tabular}
        }
        \caption{HotpotQA}
    \end{subtable}
    \begin{subtable}[t]{0.48\textwidth}
        \resizebox{\textwidth}{!}{
        \begin{tabular}{c| c c c}
        \toprule 
            \textbf{Group} &  \textbf{Factual} & \textbf{Non-Factual} & \textbf{Total}\\
            \midrule 
            log probability & $-0.127$ & $-0.220$ & $-0.172$ \\  
            Entropy & $0.163$ & $0.274$ & $0.217$ \\ 
            \bottomrule
        \end{tabular}
        }
        \caption{NQ-open}
    \end{subtable}
    \caption{Statistics of model reference. The results are broken by the factuality of the model reference.}
    \label{Table1}
\end{table*}
\begin{table}[h]
    \centering
    \begin{adjustbox}{width=.5\textwidth}
    \begin{tabular}{c|c}
    \hline 
       \textbf{ Expression Type} &  \textbf{Template}\\
    \hline 
         Uncertainty &  I am not sure but it could be\\
         Uncertainty &  I would need to double check but maybe it is \\ 
         Certainty   &  It must be \\ 
         Certainty   &  Undoubtedly it is \\ 
         \hline 
    \end{tabular}
    \end{adjustbox}
    \caption{Expressions for the prompt construction}
    \label{exp}
\end{table}
\subsection{Uncertainty Estimation for Large Language models}
The concept of uncertainty in a prediction can be thought of as the entropy of the output distribution. 
For sequence prediction tasks like natural language generation (NLG), token probabilities given past tokens have been used to represent model certainty. 
Further studies in uncertainty estimation categorize sources of uncertainty into aleatoric and epistemic uncertainty, from underlying data distribution and from missing information, respectively~\cite{kendall2017uncertainties,hullermeier2021aleatoric}. 
These uncertainties could be leveraged for decoding strategies, controllable generation, and evaluation of decisions under uncertainty ~\cite{baan2023uncertainty}.
Recently, uncertainty measures for semantics have been studied, addressing the challenges of free-form NLGs ~\cite{Kuhn2023SemanticUL,lin2023generating}.

\section{The Impact of Uncertainty and Certainty on Question Answering} 
In this section, we thoroughly analyze the behavior of LLMs under uncertain or certain expressions.
\label{preliminary}

\begin{table*}[t]
    \begin{subtable}[t]{\textwidth}
    \centering 
    \begin{adjustbox}{width=\textwidth}
    \begin{tabular}{c| c c c c c c}
    \toprule
    \textbf{Expression} & \textbf{Accuracy (\%)} & \textbf{Consistency(F, \%)} & \textbf{Consistency(NF, \%)} & $\log \frac{p_2}{p_1}$ (F) & $\log \frac{p_2}{p_1}$ (NF) & \textbf{Entropy}\\ 
    \midrule
    It must be  & $36.7$ & $82.8$ & $32.9$ & $-0.132$ & $-0.112$ & $0.421$ \\ 
    Undoubtedly it is  & $35.8$ & $90.4$ & $42.8$ & $-0.148$ & $-0.144$ & $0.438$ \\
    I am not sure but it could be & $36.3$ & $87.53$ & $45.19$ & $-0.112$ & $-0.173$ & $0.462$ \\ 
    I would need to double check but maybe it is & $35.0$ & $86.12$ & $46.3$ & $-0.193$ & $-0.212$ &  $0.528$ \\  
    \bottomrule
    \end{tabular}
    \end{adjustbox}
    \caption{HotpotQA}
    \end{subtable}
    
    \begin{subtable}[t]{\textwidth}
    \begin{adjustbox}{width=\textwidth}
    \begin{tabular}{c| c c c c c c}
    \toprule
    \textbf{Expression} & \textbf{Accuracy (\%)} & \textbf{Consistency(F, \%)} & \textbf{Consistency(NF, \%)} & $\log \frac{p_2}{p_1}$ (F) & $\log \frac{p_2}{p_1}$ (NF) & \textbf{Entropy}\\ 
    \midrule
    It must be & $58.1$ & $88.9 $ & $40.9 $ & $-0.172$ & $-0.183$ & $0.429$ \\ 
    Undoubtedly it is & $57.2 $ & $90.1 $ & $46.2 $ & $-0.182$ & $-0.184$ & $0.439$ \\ 
    I am not sure but it could be & $58.3 $ &\textbf{} $90.3 $ & $43.28 $ & $-0.149 $ & $-0.228 $ & $0.415$ \\ 
    I would need to double check but maybe it is & $54.2 $ & $88.4$ & $48.4$ & $-0.190$ & $-0.250$ & $0.458$ \\ 
    \bottomrule
    \end{tabular}
    \end{adjustbox}
    \caption{NQ-open}
    \end{subtable}
    \caption{Results of Accuracy, Consistency, Log probability ratio, and Entropy on the two benchmarks. \textbf{(F)} denotes the Factual group and \textbf{(NF)} denotes the non factual group.}
    \label{main table}
\end{table*}

\subsection{Experimental Setting}

\paragraph{Benchmark}
We evaluate \texttt{text-davinci-003} model on two closed-book QA datasets, HotpotQA \cite{Yang2018HotpotQAAD} and NQ-open \cite{kwiatkowski-etal-2019-natural}. We filtered out comparative questions such as "Who was born first, Kiefer Sutherland or Christian Slater?" since the model may answer correctly in coincidence. 
In addition, we excluded questions where the correct answer depends on the time, such as "What is the correct OS version of Apple IOS?". We use a subset of the development split of HotpotQA with $1000$ questions, and NQ-open with $997$ questions.  
\paragraph{Annotation}
First, we obtain the reference response of LLMs subject to the test of factuality. Here factuality means that the statements are grounded in valid sources like ground truth answers or external resources on the Web\footnote{If the ground truth answer contradicts to external resource in the Web, we discard the QA sample}.  We generate responses through greedy decoding with temperature=0.0.  
Obtaining the factuality label using an automatic score such as the ROUGE score~\cite{lin-2004-rouge} results in noisy labels. Therefore, we manually annotate the factuality label.

To minimize the annotation work, we utilize the Natural language inference (NLI) model. 
The NLI model \cite{he2023debertav} outputs the logit value for entailment, neutral, and contradiction classes given two inputs of sentences. 
We filtered out a pair of samples whose neutral logit class is the highest and annotated the factuality of the response separately. 
For HotpotQA, $34.5 \%$ of total samples are annotated as the factual class, and $54.43 \%$ of the NQ benchmark are annotated as the factual class. 

For the new prompt, we use the uncertainty expressions and certainty expressions in Table \ref{exp}. 
We construct the new prompt as 
Question:\{question\}.  \allowbreak Answer:\{Expression\}. 

With the new prompt of (un)certainty expression, we annotate two labels for each generated response. 
First, we check the consistency between the reference model response with the standard prompt and the response with the new prompt that includes (un)certain expressions. 
Second, the factuality of the response for the new prompt is also annotated. Annotators are guided to label two responses as consistent if two responses are "partially" consistent, even though two responses do not convey exactly the same meaning. 
\paragraph{Metric}
We evaluate the following metrics for the evaluation: 
\begin{itemize}
    \item Accuracy: The percentage of factual response with the new prompt based on the expression of (un)certainty 
    \item Consistency: We evaluate the percentage of samples that generate consistent responses compared to model references. The percentage is evaluated within each group: the factual group (\textbf{F}) and the non-factual group (\textbf{NF}). 
    \item $\log \frac{p_2}{p_1}$: First, we compute the length-normalized probability of model reference ($p_1$) and the response with the new prompt ($p_2$). Then, we evaluate the log probability ratio, which is defined as $\log \frac{p_2}{p_1}$. The ratio reflects the change in the confidence of the model response.
    \item Entropy $\mathcal{H}$: Entropy of model responses with each expression prompt is estimated using $n=10$ samples. Entropy reflects the uncertainty of LLMs. 
    
\end{itemize}
\paragraph{Statistics of Model Reference}
For HotpotQA, 346 samples out of 1000 samples are annotated as factual samples, and 260 samples are annotated as factual samples for the NQ-open benchmark. For HotpotQA, the average confidence of the LLM on the response is $-0.228$ and $-0.172$ for NQ-open. Finally, the average response entropy of HotpotQA is $0.281$ and $0.217$ for NQ-open. The statistics are shown in Table \ref{Table1}. 
\subsection{Results}

\subsubsection{Expression of Uncertainty vs Certainty}
Surprisingly, the expression of uncertainty and certainty both lead to an improvement in the accuracy of the response as shown in Table \ref{main table},  given that $34.5\%$ and $54.43$ $\%$ percent of samples are factual class in HotpotQA and NQ-open. 
We find that the expression of uncertainty does not necessarily lead to the increment of uncertainty in LLMs. For the NQ-open benchmark, the expression of certainty leads to more entropy than the expression "I am not sure but it could be". 
Moreover, the expression certainty results in the increment of entropy without the expression (Table \ref{Table1}).

\subsubsection{Consistency}
\paragraph{Consistency of LLMs under (un)certainty}
On both of the two benchmarks, there exists a clear difference between the factual group and the nonfactual group. The factual group generates consistent responses regardless of the expression used for the new prompt. Surprisingly, there is no clear difference in consistency between when a certain expression is used and when an uncertain expression is used. This result gives rise to the previous report that LLMs are poorly calibrated with respect to the expression of uncertainty or certainty 
\cite{Zhou2023NavigatingTG}.  
More than half of the nonfactual group shows inconsistency in their response. In particular, only $32.9 \%$ of the samples in the nonfactual group generate consistent responses with the expression "It must be". 
Another noticeable thing is that, for the numerical question like "How many acts is the ballet Rita Sangalli premiered in, in the year 1876?", the most of samples in the non-factual group show inconsistency. 
\paragraph{Consistency and Change in Confidence}
We break down the result of the log probability ratio into consistent and non-consistent groups. 

\begin{table}[h]
    \begin{subtable}[t]{0.5 \textwidth}
    \centering
    \adjustbox{width=\textwidth}
    {
    \begin{tabular}{c| c c}
    \toprule
         \textbf{Expression} &  \bf{$\log \frac{p_2}{p_1}$}(Con) & \bf{$\log \frac{p_2}{p_1}$} (Non-Con) \\ 
    \midrule 
    It must be    & $-0.090$  & $-0.149$ \\ 
    Undoubtedly it is & $-0.130$ & $-0.156$ \\ 
    I am not sure $\sim$    &  $-0.193$ & $-0.123$ \\ 
    I would need to double check $\sim$ & $-0.229$ & $-0.190$ \\ 
    \bottomrule
    \end{tabular}
    }
    \caption{HotpotQA}
    \end{subtable}
    \begin{subtable}[t]{0.5 \textwidth}
    \centering
    \adjustbox{width=\textwidth}
    {
    \begin{tabular}{c| c c}
    \toprule
         \textbf{Expression} &  \bf{$\log \frac{p_2}{p_1}$}(Con) & \bf{$\log \frac{p_2}{p_1}$} (Non-Con) \\ 
    \midrule 
    It must be    & $-0.167$  & $-0.175$ \\ 
    Undoubtedly it is & $-0.165$ & $-0.176$ \\ 
    I am not sure $\sim$    &  $-0.228$ & $-0.145$ \\ 
    I would need to double check $\sim$ & $-0.238$ & $-0.186$ \\ 
    \bottomrule
    \end{tabular}
    }
    \caption{NQ-open}
    \end{subtable}
    \caption{Log probability ratio results. \textbf{(Con)} denotes the consistent group and \textbf{(Non-Con)} denotes the non-consistent group.}
    \label{logprob_ratio}
\end{table}
The results are shown in Table \ref{logprob_ratio}. 
Although the proportion of consistent samples is similar when using the expression of uncertainty and certainty (Table \ref{main table}), the log probability ratio differs notably between the two expression types. For the consistent group, the usage of uncertain expression leads to more confidence drop (i.e., drop in the $\log \frac{p_2}{p_1}$ than the certainty expression). However, there is no clear difference in confidence drop in the non-consistent group. 
We speculate that the expression of uncertainty maintains a similar portion of the consistent group compared to the expression of certainty since the probability mass is distributed more \emph{evenly} to the other answer candidates. Therefore,  model reference with standard prompting which is a gold answer still remains through the greedy decoding. 

\begin{table}[h]
    \begin{subtable}[t]{0.5 \textwidth}
    \centering
    \adjustbox{width=\textwidth}
    {
    \begin{tabular}{c| c c c}
    \toprule
         \textbf{Expression} &  \textbf{Entropy}(Non-Con) & \textbf{Entropy}(Con) & $\Delta$\\ 
    \midrule 
    It must be    & $0.502$  & $0.330$ & $0.172$ \\ 
    Undoubtedly it is & $0.543$ & $0.363$ & $0.180$ \\ 
    I am not sure $\sim$    &  $0.704$ & $0.405$ & $0.299$\\ 
    I would need to double check $\sim$ & $0.715$ & $0.405$ & $0.300$ \\ 
    \bottomrule
    \end{tabular}
    }
    \caption{HotpotQA}
    \end{subtable}
    \begin{subtable}[t]{0.5 \textwidth}
    \centering
    \adjustbox{width=\textwidth}
    {
    \begin{tabular}{c| c c c}
    \toprule
         \textbf{Expression} &  \textbf{Entropy}(Non-Con) & \textbf{Entropy}(Con) & $\Delta$\\ 
    \midrule 
    It must be    & $0.537$  & $0.376$ & $0.161$ \\ 
    Undoubtedly it is & $0.549$ & $0.370$ & $0.179$ \\ 
    I am not sure $\sim$    &  $0.571$ & $0.336$ & $0.235$\\ 
    I would need to double check $\sim$ & $0.615$ & $0.396$ & $0.219$ \\ 
    \bottomrule
    \end{tabular}
    }
    \caption{NQ-open}
    \end{subtable}
    \caption{Entropy results broken by the consistency. \textbf{(Con)} denotes the consistent group and \textbf{(Non-Con)} denotes the non-consistent group. $\Delta$ denotes the difference in entropy between the two groups.}
    \label{entropy}
\end{table}

We evaluate the entropy per each expression and split down the result into the consistent and non-consistent groups as shown in Table \ref{entropy}. From the table, we can see the expression of uncertainty leads to more entropy compared to a certain one. In particular, the expression "I would need to double check $\sim$" results in the entropy of $0.715$ for the non-consistent group while $0.405$ for the consistent group. 

\begin{figure}[h]
\centering 
\includegraphics[width=\linewidth]{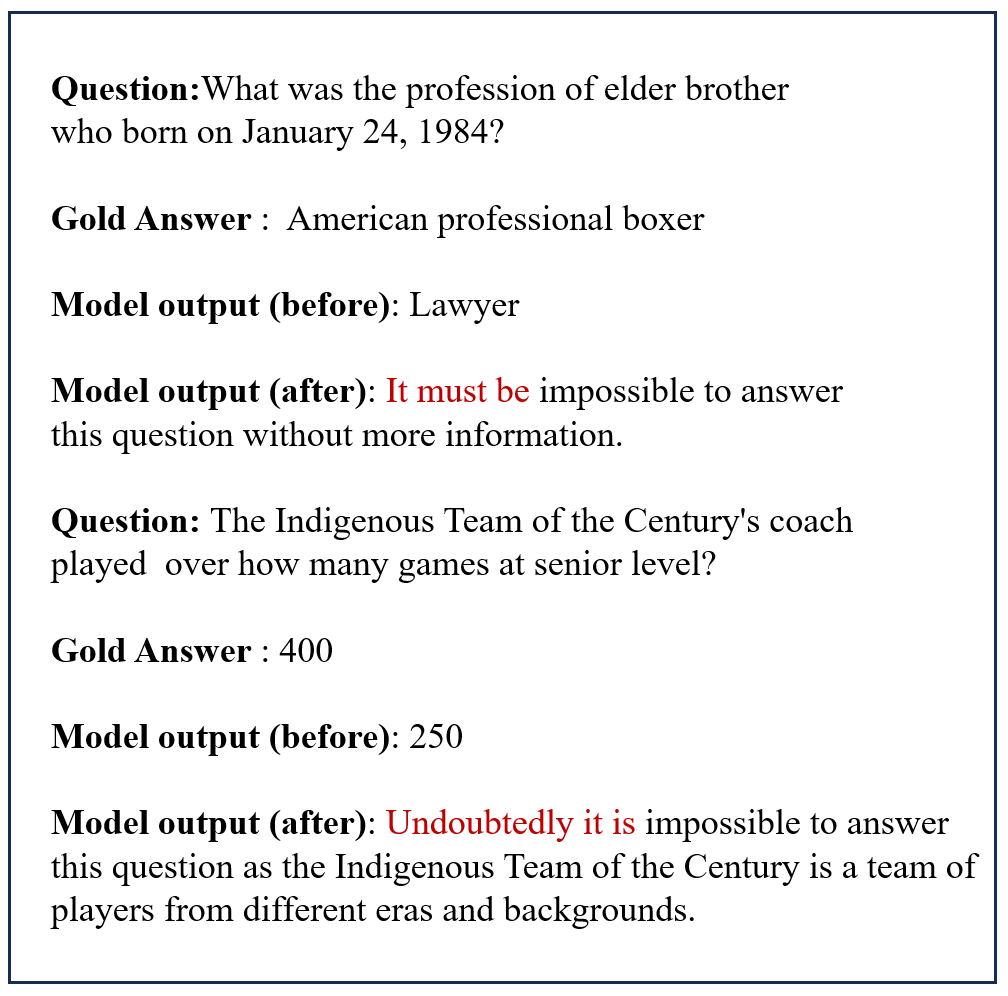}
\caption{Examples of "I can not answer" type model responses. The expressions of certainty are marked as the red color.}
\label{example-response}
\end{figure}

\paragraph{Occurrence of "I can not answer " Type Answer} 
With the standard prompting, there are few model responses that "I can not answer the answer" even if the response is not factual.
In addition, there are no responses that request the question to be more detailed. Interestingly, with an expression of certainty, some model responses fall into these types as shown in Figure \ref{example-response}. This type of response appears mainly in the expression of certainty. 

\begin{figure*}[t]
\centering
\begin{subfigure}[t]{.24\linewidth}
\includegraphics[width=\linewidth]{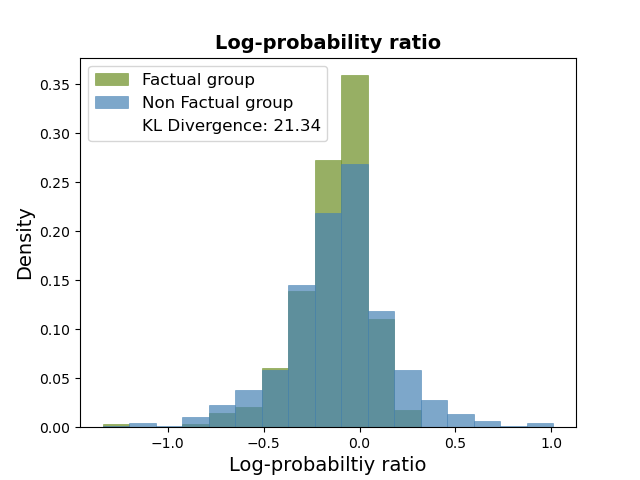}
\caption{HotpotQA: It must be \\}
\end{subfigure}
\begin{subfigure}[t]{.24 \linewidth}
\includegraphics[width=\linewidth]{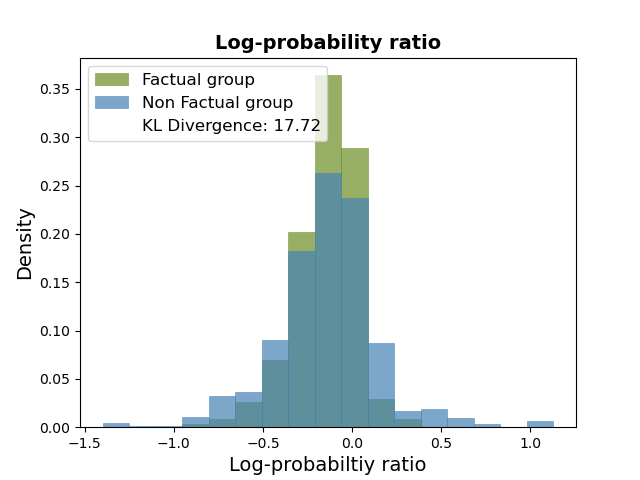}
\caption{HotpotQA: Undoutedely it is }
\end{subfigure}
\begin{subfigure}[t]{.24\linewidth}
\includegraphics[width=\linewidth]{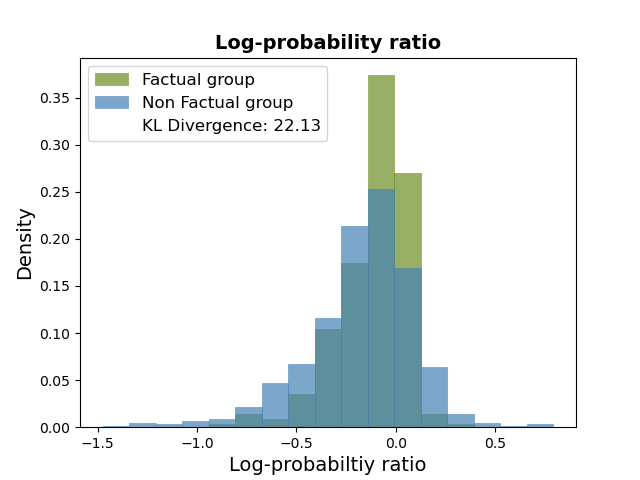}
\caption{HotpotQA: I am not sure but it could be}
\end{subfigure}
\begin{subfigure}[t]{.24\linewidth}
\includegraphics[width=\linewidth]{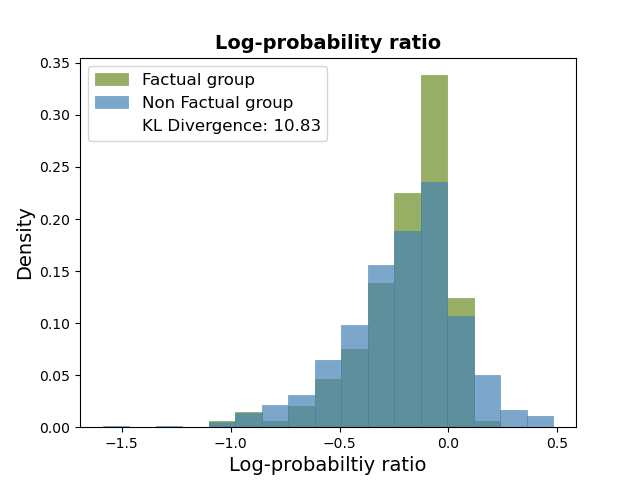}
\caption{HotpotQA: I would need to double check}
\end{subfigure}
\begin{subfigure}[t]{.24\linewidth}
\includegraphics[width=\linewidth]{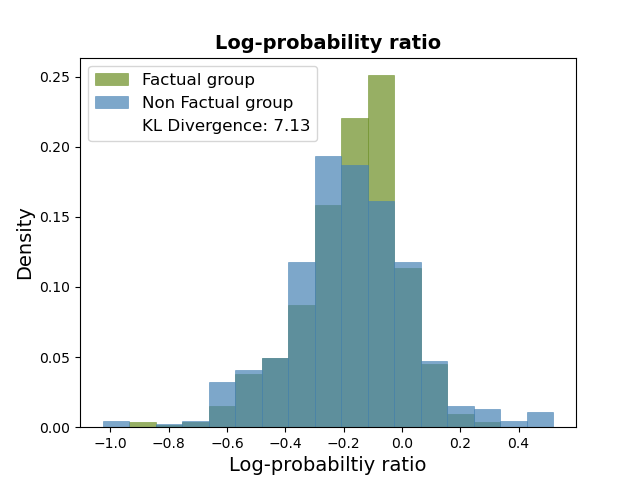}
\caption{NQ-open: It must be \\}
\end{subfigure}
\begin{subfigure}[t]{.24 \linewidth}
\includegraphics[width=\linewidth]{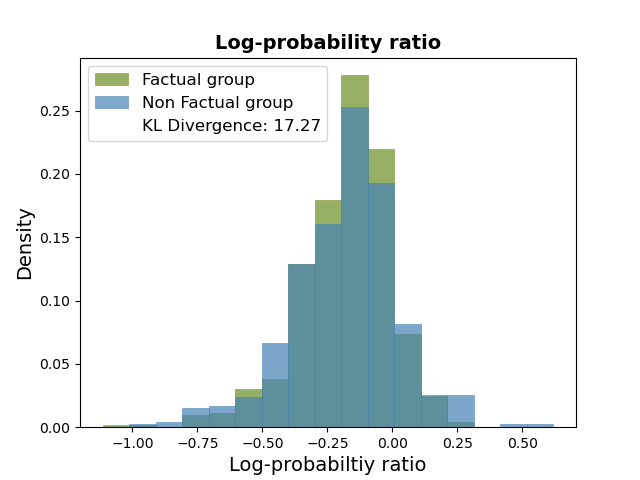}
\caption{NQ-open: Undoutedely it is }
\end{subfigure}
\begin{subfigure}[t]{.24\linewidth}
\includegraphics[width=\linewidth]{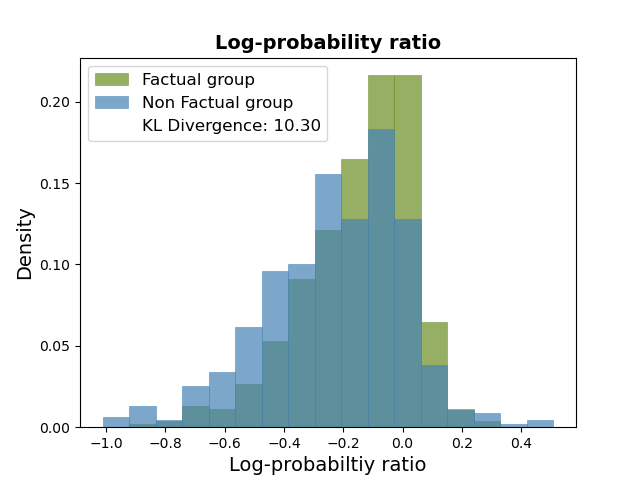}
\caption{NQ-open: I am not sure but it could be}
\end{subfigure}
\begin{subfigure}[t]{.24\linewidth}
\includegraphics[width=\linewidth]{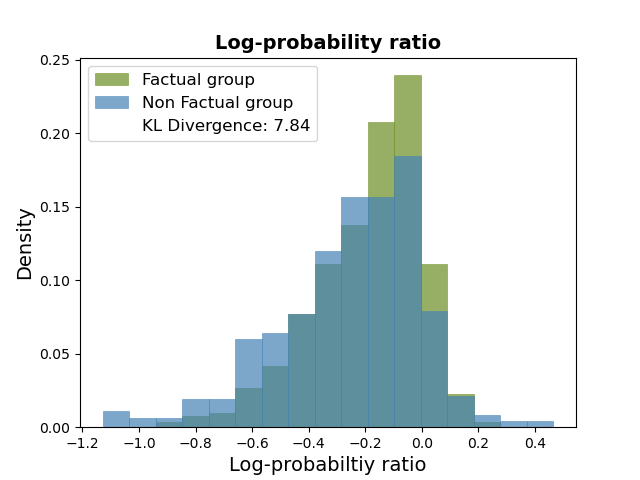}
\caption{NQ-open: I would need to double check}
\end{subfigure}
\caption{Log probability ratio distribution of non-factual group and factual group in HotpotQA and NQ-open. The distribution for the factual group is denoted as the olive color, and the distribution for the non-factual group is denoted as the blue color.}
\label{figure2}
\end{figure*}

\begin{table}[h]
    \centering
    \begin{subtable}[t]{0.4 \textwidth}
    \centering
    \adjustbox{width=\textwidth}
    {
    \begin{tabular}{c| c c }
    \toprule
         \textbf{Expression} &  $\log p_{1}$ & $\log \frac{p_2}{p_1}$\\ 
    \midrule 
    It must be    & $-0.525$  & $0.180$  \\ 
    Undoubtedly it is & $-0.578$ & $0.205$ \\ 
    \bottomrule
    \end{tabular}
    }
    \caption{HotpotQA}
    \end{subtable}
    \begin{subtable}[t]{0.4 \textwidth}
    \centering
    \adjustbox{width=\textwidth}
    {
    \begin{tabular}{c| c c }
    \toprule
         \textbf{Expression} &  $\log p_{1}$ & $\log \frac{p_2}{p_1}$\\ 
    \midrule 
    It must be    & $-0.302$  & $-0.09$  \\ 
    Undoubtedly it is & $-0.298$ & $-0.102$ \\ 
    \bottomrule
    \end{tabular}
    }
    \caption{NQ-open}
    \end{subtable}
    \caption{Average log probability and log probability ratio in the "I can not answer " type model responses.}
    \label{table6}
\end{table}
We find interesting characteristics of this type of answer. As shown in Table \ref{table6}, the log probability of model reference in these samples is notably high compared to the average. In addition, the log probability after inserting the expression does not change much and even increases. Especially, for the expression "Undoubtedly it is", the average log probability results in $0.205$ in the HoptotQA benchmark. These findings indicate that LLMs are encouraged to generate "I can not answer  the answer" under some expression when they are unconfident in answering the question. 

\subsubsection{Factuality and Confidence of LLMs}
To study the relation between the factuality and confidence of LLMs, we evaluate the log probability ratio of each sample in the factual and nonfactual groups. The ratio for each group is shown in Table \ref{table4} and the distribution is described Figure \ref{figure2}. To quantify the distribution shift, we evaluate the KL divergence between two histograms with the Gaussian smoothing filter.

For both uncertain and certain expressions, the overall probability of the responses with the new prompt decreases significantly.  One notable thing is that the use of certainty makes a similar value of log probability ratio between the factual group and the nonfactual group. However, the expression of uncertainty leads to more drop in log probability in the nonfactual group compared to the factual group. This indicates that some examples of nonfactual groups behave more sensitively under the expression of uncertainty. 
\vskip -0.05in  
\begin{table}[h]
\begin{subtable}[t]{0.5\textwidth}
\centering
\adjustbox{width=\textwidth}
{
\begin{tabular}{c|c c c}
\toprule
\textbf{Expression} & $\log \frac{p_2}{p_1}$ (F) & $\log \frac{p_2}{p_1}$ (NF) & $\Delta_{ratio}$   
\\ \midrule 
It must be         & $-0.132$ & $-0.112$ & $0.02$       \\
Undoubtedly it is    & $-0.148$ & $-0.144$ &  $0.004$    \\
I am not sure $\sim$     & $-0.112$ & $-0.173$ & $-0.061$       \\ 
I would need to $\sim$  & $-0.193$ & $-0.212$ & $-0.019$  \\ 
\bottomrule
\end{tabular}
}
\caption{HotpotQA}
\end{subtable}
\\ 
\begin{subtable}[t]{0.5 \textwidth}
\centering
\adjustbox{width=\textwidth}
{
\begin{tabular}{c|c c c c}
\toprule
\textbf{Expression} & $\log \frac{p_2}{p_1}$ (F) & $\log \frac{p_2}{p_1}$ (NF) & $\Delta_{ratio}$      
\\ \midrule 
It must be    &  $-0.172$ & $-0.183$ & $-0.011$ \\ 
Undoubtedly it is    & $-0.182$ & $-0.184$ &  $-0.002$ \\
I am not sure $\sim$ &  $-0.149$ & $-0.228$ & $-0.079$ \\ 
I would need to $\sim$  & $-0.190$ & $-0.250$ & $-0.09$ \\ 
\bottomrule
\end{tabular}
}
\caption{NQ-open}
\end{subtable}
\caption{Statistics for the log probability ratio. $\log  \frac{p_2}{p_1}$ denotes the average log probability ratio of the Factual (F) or nonfactual (NF) group. $\Delta_{ratio}$ denotes the difference between the log probability ratio between the factual group and the nonfactual group.} 
\label{table4}
\end{table}

\begin{figure*}[t]
    \centering
    \begin{subfigure}[t]{0.4 \linewidth}
        \includegraphics[width=\linewidth]{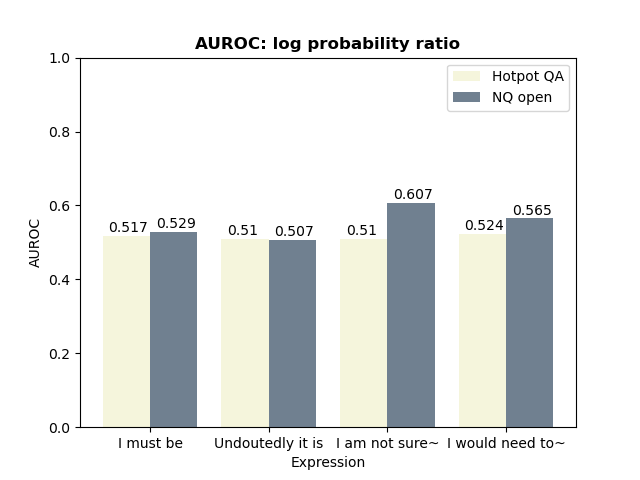}
    \caption{log probablity ratio}
    \end{subfigure}
    \begin{subfigure}[t]{0.4 \linewidth}
        \includegraphics[width=\linewidth]{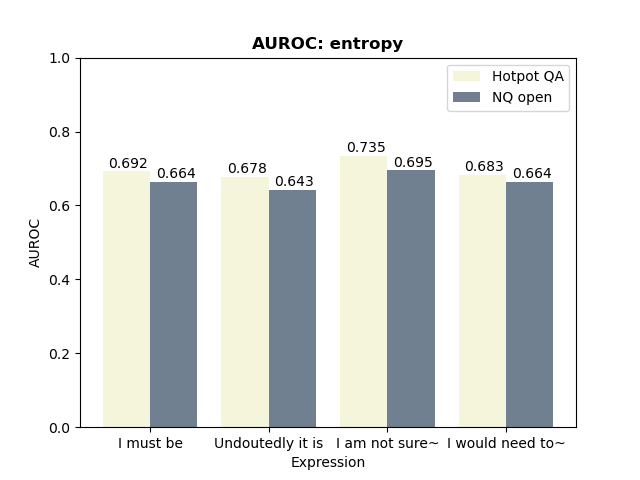}
    \caption{Entropy}
    \end{subfigure}
    \caption{AUROC bar graph for HotpotQA and NQ-open. The AUROC metric per the expression is plotted.}
    \label{auroc1}
\end{figure*}

\subsubsection{Can we detect confidently nonfactual samples?}
Although the majority of the nonfactual group changes their responses with the new prompt of uncertainty or certainty expression, there still exists a large portion of consistently wrong samples. 
Does this consistent nonfactual group have distinct features from the non-consistent \& nonfactual group? To investigate this matter, we exploit the log probability ratio as the metric for detecting nonfactual classes among the consistent group and compute the area under the receiver operator characteristic curve (AUROC) (Figure \ref{auroc1}). 
Mostly the AUROC for the metric log probability ratio remains around $0.5$ with the exception of $0.607$. This means that the nonfactual samples among consistent groups exhibit similar confidence changes compared to the factual samples, which also indicates that many nonfactual samples are \emph{confidently} generated. 
Meanwhile, the AUROC for the entropy exhibits high values compared to the previous one. This indicates that the uncertainty of LLMs may be more effective for detecting the nonfactual responses than the confidnce of LLMs. 

\section{Black-box Hallucination Detection Metric}
\label{detection metric}
Motivated by the observations in the previous section, we present a simple detection metric using the consistency of model response. First, we generate the new response using the template \texttt{Question:\{question\}. Answer:\{Expression\}} like the previous section. 
In the next step, using the single NLI model \cite{he2023debertav}, we obtain logit values for three classes (entailment, neutral, contradiction) with the input of two responses: model reference $r$ and model response with the new prompt. The input for NLI model constructed as \texttt{ \{Question\}\{Model reference\}[SEP]\{Question\}\{Model response\}}.
Given model reference $r$, we use the difference between the logit for "entailment" and "contradiction" class as the metric: 
\begin{equation*}
    f_{score} (r) = logit_{\text{contradiction}} - logit_{\text{entailment}}.  
\end{equation*}
As a variant, we adopt ensemble methods. Specifically, We take the minimum of the score using the two expressions (e.g., "I am not sure $\sim$" and "It must be"). 
\begin{equation*}
    f_{score2} (r) = \min \{f_{score, \; exp1}, f_{score, \; exp2}\}
\end{equation*}
Given that there are many consistently wrong samples, this variant may boost the detection performance.
\section{Hallucination Detection Experiments}
In this section, we evaluate the performance of our detection metric on the full development split of HopotQA with 5949 questions and NQ-open with 3610 questions. 
\label{section4}
\subsection{Baseline}
We compare our metrics against the baseline related to the uncertainty or confidence of LLMs including log probability of the model response, entropy, semantic entropy, semantic similarity, and SelfCheckGPT in \citet{Manakul2023SelfCheckGPTZB}. \textbf{Semantic entropy} \footnote{\url{https://github.com/lorenzkuhn/semantic_uncertainty}} uses the semantic equivalence between the generated responses. First, generate $m$ samples of responses to estimate the entropy. 
Next, we group semantically equivalent responses using the NLI model. 
For example, given the question "Where is the capital of France?", the responses "It's Paris" and "Paris, I think" are grouped together. Finally, for the generated group $C_1, \dots, C_n$, compute the semantic entropy by merging the probability mass in the same group: 
\begin{align*}
    \textbf{SE} &= - \sum_{i=1}^n p(c_i) \cdot \log p (c_i) \\&= - \sum_{i=1}^n \left( (\sum_{s \in c_i} p(s | x)) \log(\sum_{s \in c_i} p (s | x)) \right).  
\end{align*}
\textbf{Lexical similarity} uses the consistency of the generated responses of LLMs. Generate $m$ response, then compute the lexical similarity between samples:  
\begin{equation*}
    \textbf{LS} = \frac{1}{m(m-1)} \sum_{i=1}^m \sum_{n \neq m} \text{sim} (r_m, r_n).  
\end{equation*}

\begin{figure*}[t]
\begin{subfigure}[t]{0.45 \linewidth}
\includegraphics[width=\linewidth]{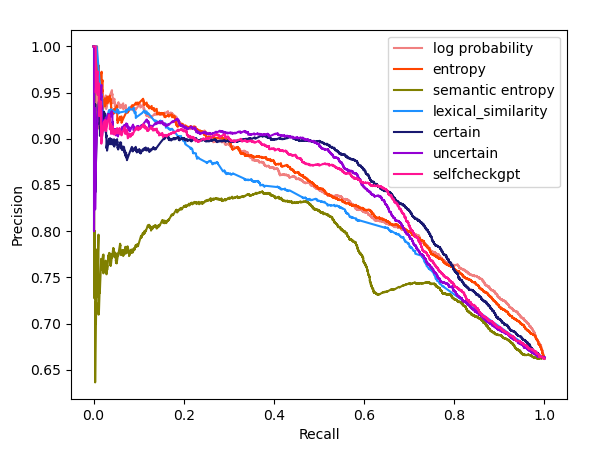}
\caption{HotpotQA}
\end{subfigure}
\begin{subfigure}[t]{0.45 \linewidth}
\includegraphics[width = \linewidth]{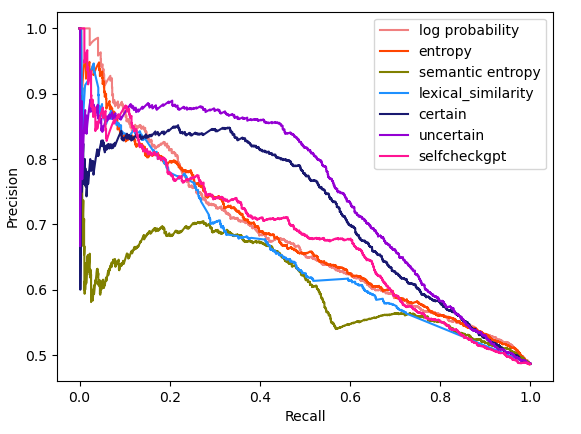}
\caption{NQ-open}
\end{subfigure}
\caption{PR-Curve of detecting and factual and non-factual respones of GPT-3 in HotpotQA and NQ-open.}
\label{PRcurve}
\end{figure*}

We use rouge-L metric for evaluating the lexical similarity between the responses. For the SelfCheckGPT baseline, we choose the NLI version of the model since it shows better performance compared to other SelfCheckGPT versions including BERTscore, unigram probability, and question answering. The probability of the NLI model for the "contradiction" class is utilized as the metric with the input of model reference and sampled answer\footnote{We use the number of samples $n=8$ and sampling temperature of $\tau=0.5$.}. 
\subsection{Evaluation Metric}
We use two metrics to evaluate the performance of the metric: Area Under the Receiver Operator Characteristic curve (AUROC) and Area Under the Precision-Recall Curve (AUPRC). In both metrics, being closer to the value 1.0 indicates that the metric is a better predictor of factuality.
\subsection{Results}
\begin{table}[h]
    \begin{subtable}[t]{0.5 \textwidth}
    \centering
    \begin{tabular}{c| c c}
        \toprule 
        \textbf{Method} &  \textbf{AUROC} & \textbf{AUPRC} \\ \midrule
         $\log p$ & $0.739$ &  $0.838$ \\ 
         Entropy  & $0.735$ &  $0.838$ \\ 
         Semantic entropy & $0.665$ & $0.774$ \\    
         Lexical similarity & $0.700$ & $0.810$ \\
         SelfCheckGPT(NLI)   & $0.730$  & $0.837$ \\ 
         \midrule   
         $f_{certain}$    & $\textbf{0.746}$ & $\textbf{0.842}$ \\ 
         $f_{uncertain}$  & $0.727$ & $0.839$ \\ 
         $f_{ensemble}$   & $0.717$ & $0.836$  \\ \bottomrule
    \end{tabular}
    \caption{HotpotQA}
    \end{subtable}
    
    \begin{subtable}[t]{0.5 \textwidth}
    \centering
    \begin{tabular}{c| c c}
    \toprule 
        \textbf{Method} &  \textbf{AUROC} & \textbf{AUPRC} \\ \midrule
        $\log p$ & $0.680$ & $0.680$ \\ 
        Entropy & $0.681$ & $0.677$ \\ 
        Semantic entropy & $0.638$ & $0.608$ \\ 
        Lexical similarity &  $0.651$   & $0.647$ \\ 
        SelfCheckGPT(NLI)   &  $0.681$   & $0.681$ \\ 
        \midrule  
        $f_{certain}$   &  $0.723$ & $0.718$ \\ 
        $f_{unceratin}$ &  $\textbf{0.741}$ & $\textbf{0.748}$ \\ 
        $f_{ensemble}$  &  $0.726$ & $0.739$ \\ \bottomrule
    \end{tabular}
    \caption{NQ-open}
    \end{subtable}
    \caption{AUROC and AUPRC evaluation of the metrics. $f_{certain}$ denotes our metric using the certain expression: "It must be", $f_{unceratin}$ denotes our metric using the uncertain expression: "I am not sure but it could be", and $f_{ensemble}$ denotes the ensemble method.}
    \label{metric-result}
    \vspace{-.3in}
\end{table}

The AUROC and AUPRC evaluation results are shown in Table \ref{metric-result}. For both HotpotQA and NQ-open, our metric outperforms the baselines in predicting whether the response of LLMs to a question are factual.  In HotpotQA, our metric with the expression "It must be" demonstrates the best AUROC and AUPRC. Meanwhile, in NQ-open benchmark, our metric with the uncertain expression "I am not sure but it could be" shows the best performance. Notably, methods using the token probability show worse performance compared to HotpotQA,  which indicates that the LLM is relatively not well calibrated with NQ-open. 

\section{Discussion and Conclusion}

In this paper, we conduct a comprehensive analysis of the behavior of GPT-3 \texttt{text-davinci-003} under expression of uncertainty and certainty and propose a simple black-box hallucination detection metric based on the expression of uncertainty. The primary finding is that the LLM behaves differently under the expression when it generates a factual response and a nonfactual response. In addition, the LLM is \emph{not} linguistically calibrated with respect to the expression of certainty. The certain expression does not improve the accuracy of the response, and it leads to a decrease in the log probability of the response.

Most samples in the factual group show consistency and 
the majority in the nonfactual group show inconsistency when the new prompt of (un)certainty is used. 
However, there still exists some sample in the nonfactual group that consistently gives the nonfactual response. This kind of sample makes the hallucination task more challenging since these samples have no distinct feature compared to the factual response. For more successful hallucination detection, we should train the LLM in a manner that assigns low probabilities to non-factual statements.

A critical challenge revealed by our analysis is that LLMs often assign high confidence to responses that are not factually grounded. This occurs because many statements that are \textit{plausible but incorrect} still receive high probability under the model’s distribution. To mitigate this issue, a promising direction is to apply \textit{unlikelihood training}~\citep{welleck2020unlikelihood} as a corrective mechanism. The key idea is to identify nonfactual statements that the LLM tends to generate with high confidence—for example, statements involving entities that are closely related in semantic or relational space—and then penalize these outputs through unlikelihood training. By explicitly discouraging the model from overconfidently producing such plausible but nonfactual responses, this approach can improve factual calibration and contribute to more robust hallucination mitigation strategies.

\newpage 
\section{Bibliographical References}\label{sec:reference}
\bibliographystyle{lrec-coling2024-natbib}
\bibliography{reference}

\begin{table*}[ht]
\centering
\small
\renewcommand{\arraystretch}{1.3}
\setlength{\tabcolsep}{6pt}

\begin{subtable}[t]{\textwidth}
\centering 
\begin{tabular}{p{0.28\textwidth} p{0.15\textwidth} | p{0.18\textwidth} p{0.28\textwidth} c}
\toprule
\textbf{Question} & \textbf{Ground truth} & \textbf{Normal prompt} & \textbf{(Un)certain prompt} & $f_{score}$ \\
\midrule
When was the plaintiff in the 1892 Barbed Wire Patent Case born? &
January 18, 1813 &  
1839 & 
\textcolor{blue}{I am not sure} but it could be 1845 & 
7.64 \\
\midrule
What is the name of the American author to Grace Nail Johnson who was the first African American executive secretary of NAACP? & 
James Weldon Johnson & 
W.E.B. Du Bois & 
\textcolor{blue}{I am not sure} but it could be Walter White. & 
7.83 \\
\midrule
Which star of Zork was also the voice of Pac-Man? & 
Martin Ingerman & 
Ernie Anderson & 
\textcolor{red}{It must be} Peter Cullen, who voiced Pac-Man in the 1982 animated series & 
8.08 \\
\midrule
Who was the wife of the United States Army lieutenant general who received the Distinguished Service Cross? & 
Julia Compton Moore & 
Mary Louise Hines & 
\textcolor{red}{It must be} Mamie Eisenhower & 
6.37 \\
\bottomrule
\end{tabular}
\caption{Detected as hallucinated samples with a threshold=0.5 (after normalization).}
\end{subtable}

\vspace{0.8em}

\begin{subtable}[t]{\textwidth}
\centering 
\begin{tabular}{p{0.28\textwidth} p{0.15\textwidth} | p{0.18\textwidth} p{0.28\textwidth} c}
\toprule
\textbf{Question} & \textbf{Ground truth} & \textbf{Normal prompt} & \textbf{(Un)certain prompt} & $f_{score}$ \\
\midrule
What is 25 miles south of Groom Lake? & 
Rachel, Nevada &  
Rachel, NV & 
\textcolor{blue}{I am not sure} but it could be Las Vegas, Nevada. & 
-6.70 \\
\midrule
When was Leeds no longer called Elmet? & 
early 7th century & 
7th century AD & 
\textcolor{blue}{I am not sure} but it could be around the 7th century. & 
-4.41 \\
\midrule
Sporobolus and Zea are in the same what? & 
family & 
Family &  
\textcolor{blue}{I am not sure} but it could be family or genus. & 
-7.32 \\
\midrule
What is Opry Mills in Nashville, Tennessee? & 
super-regional shopping mall & 
Opry Mills is a large shopping mall in Nashville, Tennessee & 
\textcolor{red}{It must be} a shopping mall. & 
-7.60 \\
\bottomrule
\end{tabular}
\caption{Detected as non-hallucinated samples with a threshold=0.5 (after normalization).}
\end{subtable}

\caption{Examples of hallucination detection. Blue highlights indicate uncertain responses (\textit{I am not sure}), while red highlights indicate overconfident responses (\textit{It must be}).}
\label{main table}
\end{table*}

\end{document}